\documentclass[10pt,twocolumn,letterpaper]{article}

\usepackage{cvpr}
\usepackage{times}
\usepackage{epsfig}
\usepackage{graphicx}
\usepackage{amsmath}
\usepackage{amssymb}

\usepackage{listings}

\usepackage{xcolor}

\usepackage{xcolor}
\definecolor{cgreen}{RGB}{34, 139, 34}
\usepackage{pifont}
\newcommand{\cmark}{{\color{cgreen}\ding{51}}}%
\newcommand{\xmark}{{\color{red}\ding{55}}}%

\definecolor{cgreen}{RGB}{34, 139, 34}
\usepackage{pifont}

\usepackage[pagebackref=true,breaklinks=true,letterpaper=true,colorlinks,bookmarks=false]{hyperref}

\cvprfinalcopy 

\def\cvprPaperID{****} 

\ifcvprfinal\fi

\begin{document}

\title{SIDOD: A Synthetic Image Dataset for \\ 3D Object Pose Recognition with Distractors}

\author{Mona Jalal\\
Boston University\\
 {\tt\small jalal@bu.edu}
\and
Josef Spjut\\
NVIDIA\\
 {\tt\small jspjut@nvidia.com}
\and
Ben Boudaoud\\
NVIDIA\\
 {\tt\small bboudaoud@nvidia.com}
\and
Margrit Betke\\
Boston University\\
 {\tt\small betke@bu.edu}
}

\maketitle

\begin{abstract}
    We present a new, publicly-available image dataset generated by the NVIDIA Deep Learning Data Synthesizer intended for use in object detection, pose estimation, and tracking applications. This dataset contains 144k stereo image pairs that synthetically combine 18 camera viewpoints of three photorealistic virtual environments with up to 10 objects (chosen randomly from the 21 object models of the YCB dataset~\cite{calli2015ycb}) and flying distractors.  Object and camera pose, scene lighting, and quantity of objects and distractors were randomized.   
     Each provided view includes RGB, depth, segmentation, and surface normal images, all pixel level.
    We describe our approach for domain randomization and provide insight into the decisions that produced the dataset.
\end{abstract}


\vspace*{-0.7cm}

\section{Introduction}
Computer vision has shifted towards the use of neural networks, trained on large datasets, to exceed the performance of more traditional algorithms in object recognition, pose estimation, and object tracking.
In order for training to be effective, the training set needs to exhibit similar characteristics to the target application, while maintaining enough variation to prevent overfitting.
While a number of datasets to support object recognition research already exist, which provide a variety of labels~\cite{hodan2017t,lai2011large,rennie2016dataset,tremblay2018falling}, there is still a need for larger, high-quality datasets.
To this end, we present a new synthetic dataset called {\sc SIDOD}\footnote{\url{https://research.nvidia.com/publication/2019-06_SIDOD\%3A-A-Synthetic}} as an acronym for Synthetic Image Dataset for 3D Object Pose Recognition with Distractors present.
We are motivated by the notion that object pose and segmentation are much easier to capture with synthetic tools rather than custom-built sensors.
An added benefit is other potentially useful or interesting pieces of information, such as surface normals, are available in the rendering process used to generate synthetic images.

Given that {\sc SIDOD} provides 3D pose and rotation along with depth and pixel-level segmentation, it may be useful for a number of applications in robotics and computer vision.
We were motivated by potential uses in Virtual and Augmented Reality (VR/AR) where object understanding and tracking can be vital to immersive user experiences.
One challenge in tracking hand-held objects is the necessary occlusion of the user's hand, over a range of grasps, as well as a wide range of possible colors and textures.
To provide random occlusion to approximate these anticipated situations, we added subsets of images with \emph{flying distractors}~\cite{to2018ndds} to our dataset. 
\emph{Flying distractors} are 3D shapes, which are not objects of interest.  


\begin{figure}
    \centering
    \includegraphics[width=0.485\columnwidth]{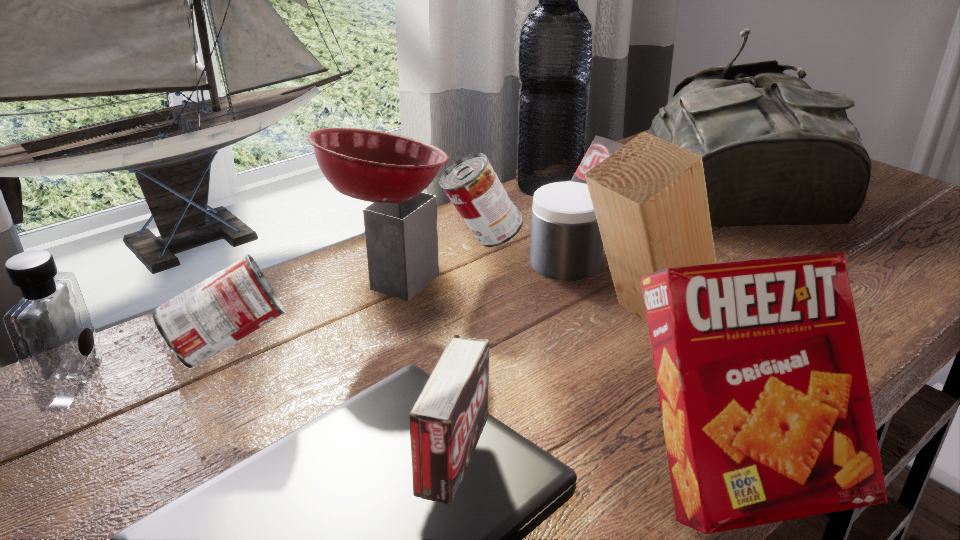}
    \includegraphics[width=0.485\columnwidth]{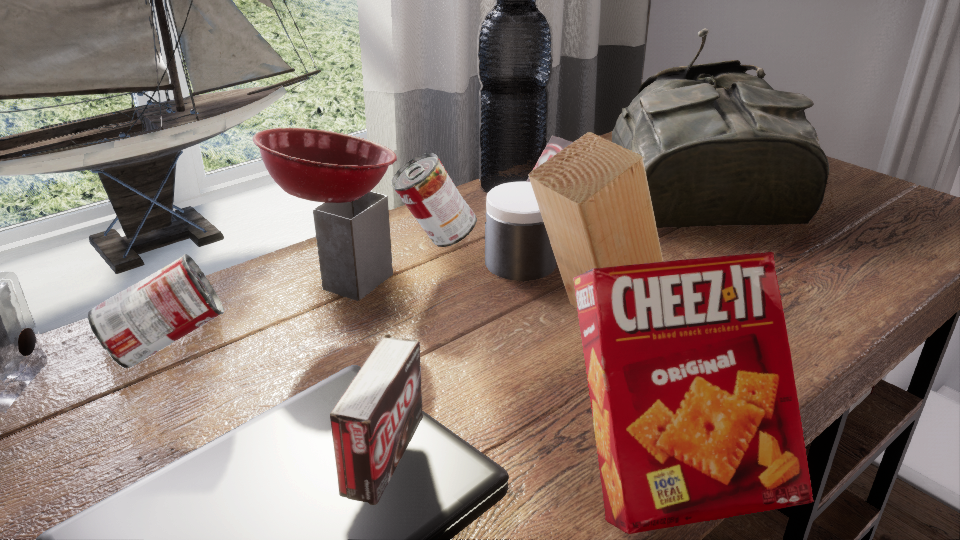}    \includegraphics[width=0.32\columnwidth]{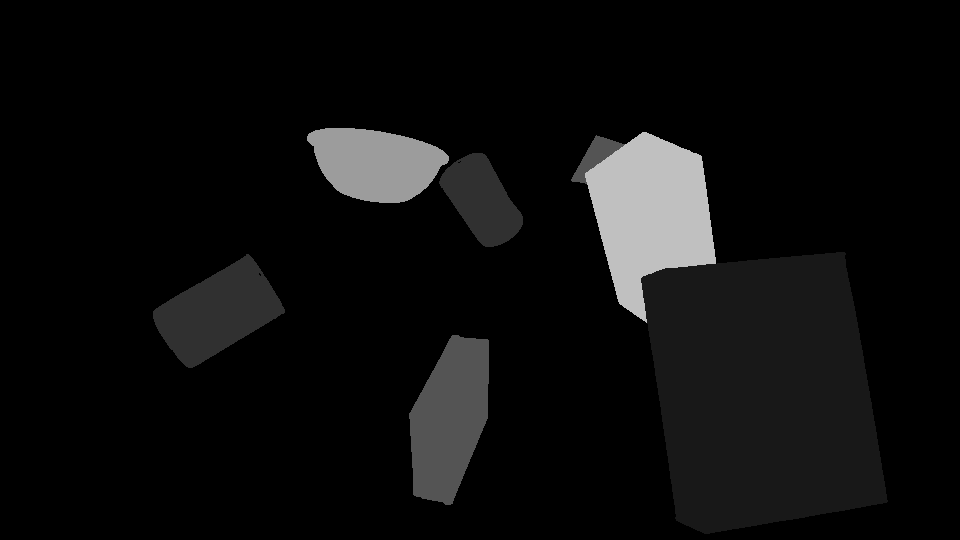}
    \includegraphics[width=0.32\columnwidth]{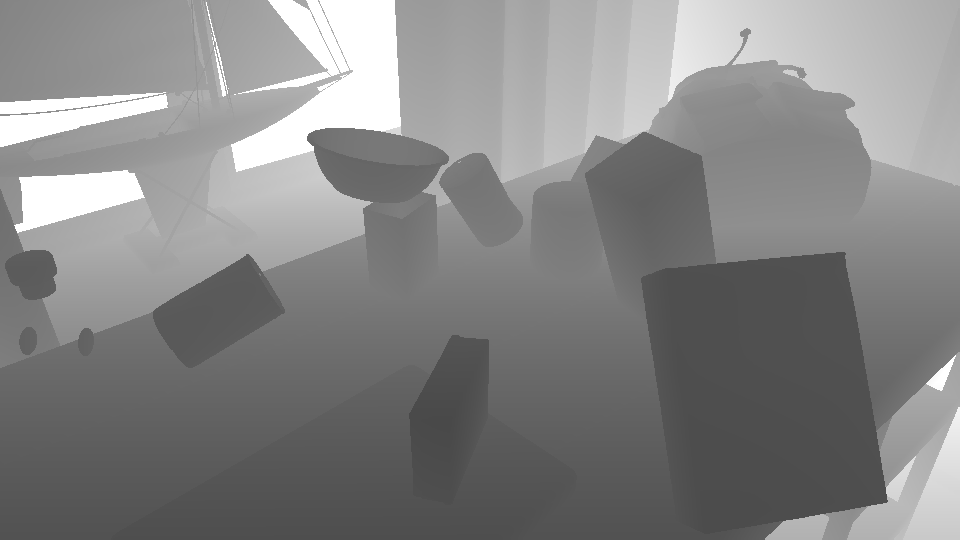}
    \includegraphics[width=0.32\columnwidth]{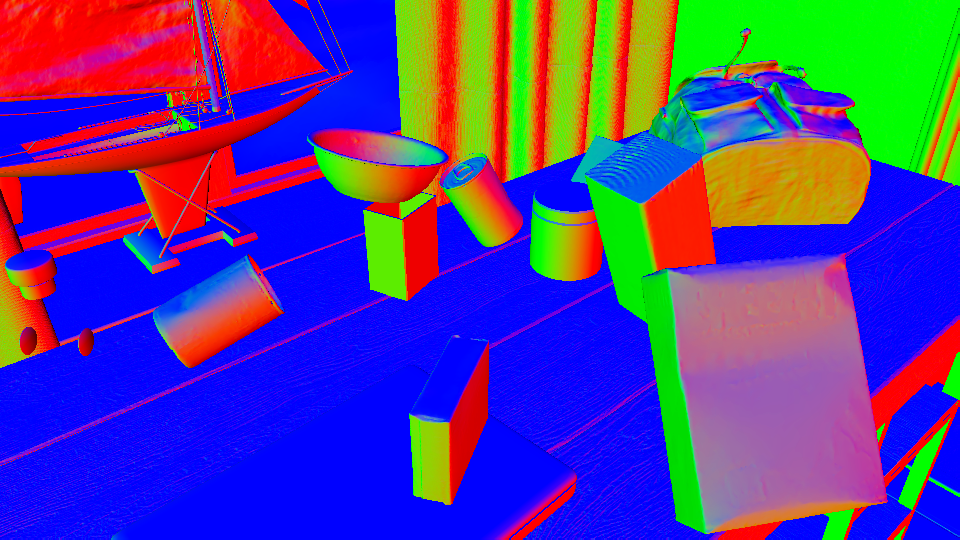}
    \caption{Examples from the dataset. Left and right stereo views (top) with pixel-level segmentation, depth, and surface normals for the left view (bottom, from left to right). 
    \vspace*{-0.5cm}}
    \label{fig:example}
\end{figure}

\begin{figure}
    \centering
    \includegraphics[width=0.485\columnwidth]{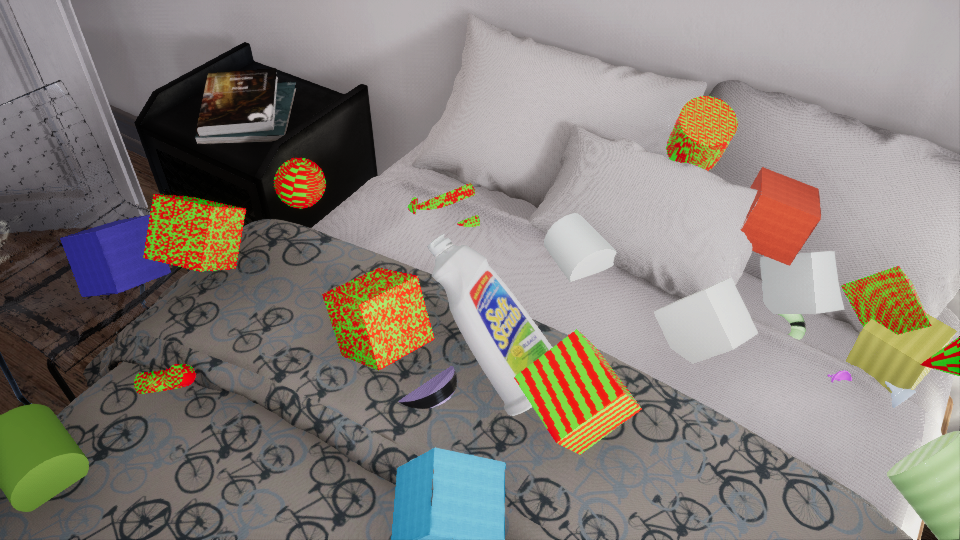}
    \includegraphics[width=0.485\columnwidth]{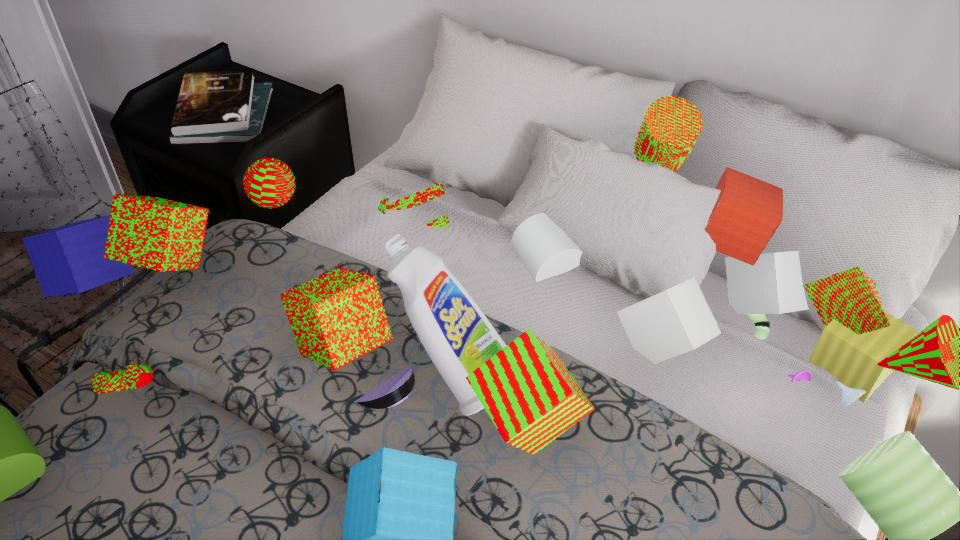} \includegraphics[width=0.32\columnwidth]{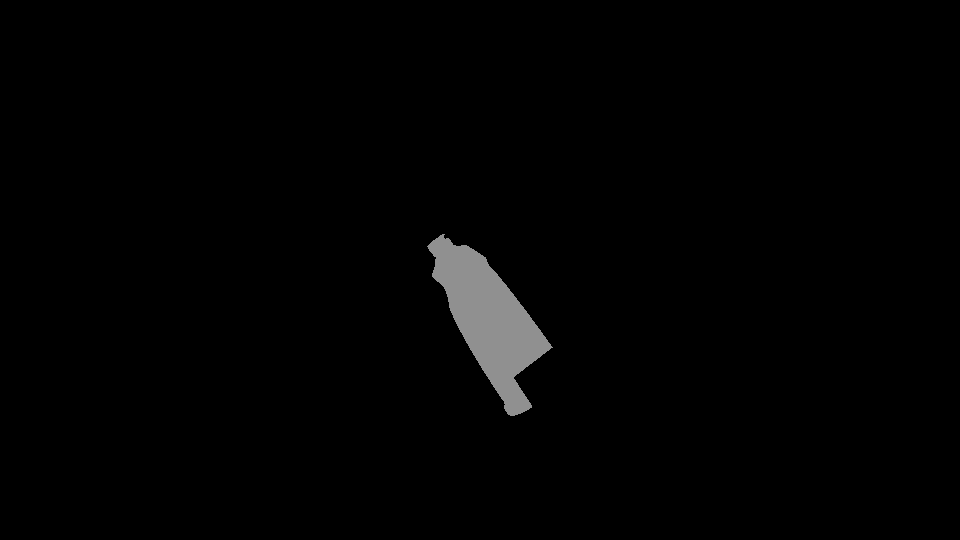} \includegraphics[width=0.32\columnwidth]{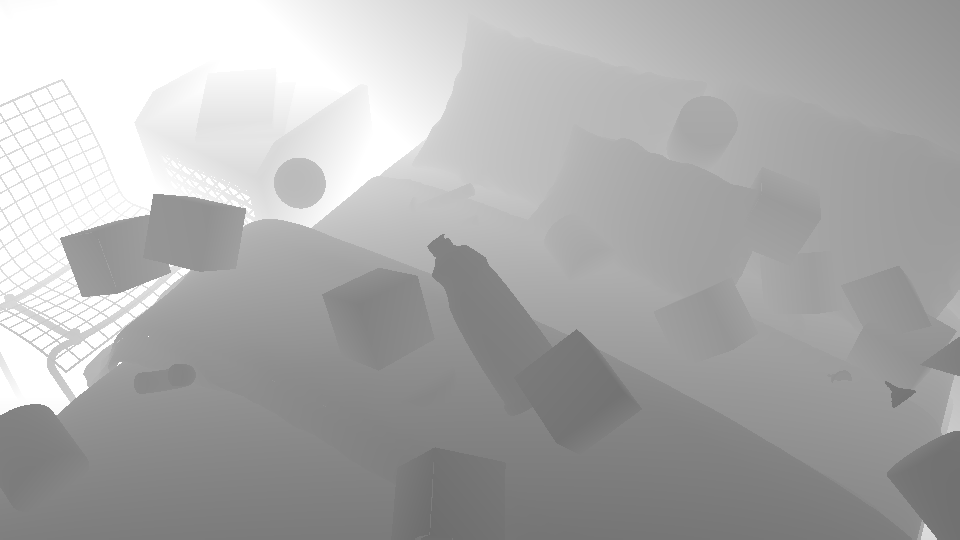} \includegraphics[width=0.32\columnwidth]{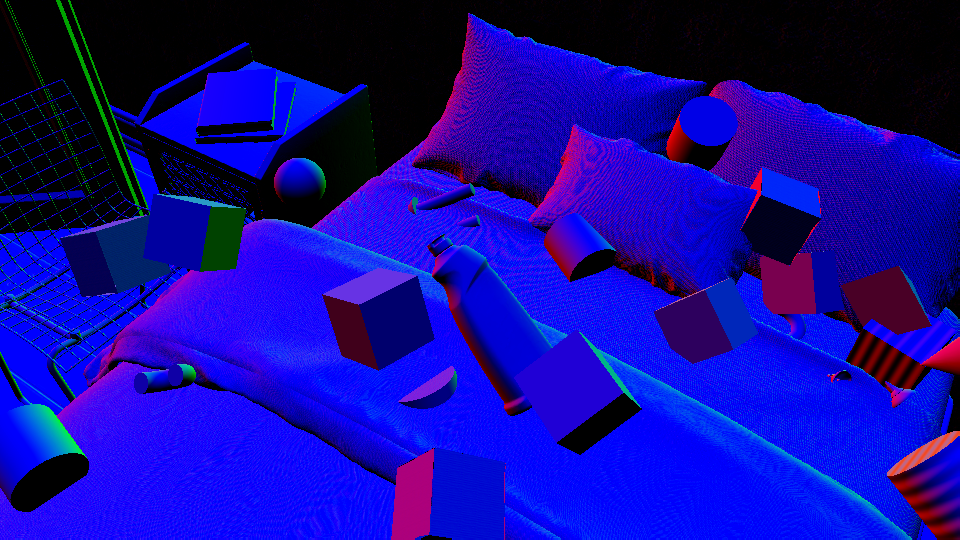}
    \caption{Single tracked object (bleach cleanser) with flying distractors.
    \vspace*{-0.3cm}}
    \label{fig:distractor}
\end{figure}

Similarly to the {\sc Falling Things} dataset, we provide stereo views as RGB, depth, and pixel-level segmentation images and increase the amount of available data with significant object occlusion. 
Additionally, we include flying distractors in a subset of views, representing even more aggressive occlusion conditions (Fig.~\ref{fig:distractor}). We also provide pixel-level surface normals for all view pairs. A comparison with existing datasets is given in Table~\ref{tab:comparison}.

\newcommand{\rb}[1]{\makebox[20pt]{\rotatebox{45}{#1}}}%
\begin{table*}[ht]
    \begin{tabular}{rrrc|ccccccccc}
    Dataset & \#objects  &  \#frames & Description & 
    \makebox[10px]{\rb{depth}} & \makebox[10px]{\rb{stereo}} & 
    \makebox[9px]{\rb{\parbox{0.7cm}{\centering 3D \\ pose}}}  &
   \rb{\parbox{1.1cm}{\centering full \\ rotation}}  & \rb{occlusion} & 
   \rb{\parbox{1.2cm}{\centering extreme \\ lighting}}  & 
    \rb{\parbox{1.2cm}{\centering segmen \\ tation}}  & 
    \rb{\parbox{0.9cm}{\centering bbox \\ coords}} & 
    \rb{\parbox{1.5cm}{\centering flying \\ distractors}} \\
    \hline

    UW RGB~\cite{lai2011large} & 300 & 250k & various & 
    \cmark & \xmark & \cmark & \cmark & \xmark & \xmark & \xmark & \xmark & \xmark \\

    LINEMOD~\cite{hinterstoisser2012model} & 15 & 18k & various & 
    \cmark & \xmark & \cmark & \cmark & \xmark & \xmark & \xmark & \xmark & \xmark \\

    Pascal3D+~\cite{xiang2014beyond} & 12 & 30k & various & 
    \xmark & \xmark & \cmark & \xmark & \cmark & \cmark & \xmark & \xmark & \xmark \\


    Rutgers APC~\cite{rennie2016dataset} & 24 & 10k & warehouse & 
    \cmark & \xmark & \cmark & \xmark & \xmark & \xmark & \xmark & \xmark & \xmark \\

    T-LESS~\cite{hodan2017t} & 30 & 10k & industrial & 
    \cmark & \xmark & \cmark & \cmark & \cmark & \xmark & \xmark & \xmark & \xmark \\

    YCB Video~\cite{xiang2017posecnn} & 21 & 134k & household & 
    \cmark & \xmark & \cmark & \cmark & \cmark & \xmark & \cmark & \cmark & \xmark \\

    Falling Things~\cite{tremblay2018falling} & 21 & 60k & household & 
    \cmark & \cmark & \cmark & \cmark & \cmark & \cmark & \cmark & \cmark & \xmark \\

    SIDOD & 21 & 144k & household & 
    \cmark & \cmark & \cmark & \cmark & \cmark & \cmark & \cmark & \cmark & \cmark \\
         & 
    \end{tabular}
    \caption{Object pose and detection datasets with at least 10,000 frames.} 
    \label{tab:comparison}
\end{table*}

\section{SIDOD Dataset}


We generated our dataset using the NVIDIA Deep Learning Data Synthesizer (NDDS~\cite{to2018ndds}). Built on top of Unreal Engine 4.18.3, this tool captures view data at a much higher rate than many other available rendering solutions, as high as tens of frames per second.
We selected three distinct virtual environments and manually chose a variety of distinct locations within each environment to provide sufficient variations of viewpoints to mimic a variety of indoor locations. The scenes are as follows:
bakery (4 locations), AIUE{\textunderscore}V02{\textunderscore}001 (10 locations), AIUE{\textunderscore}V02{\textunderscore}002 (4 locations) for a total of 18 different locations.
Corresponding directories can be found for each location. The scenes prefixed with AIUE\footnote{\url{https://evermotion.org}} represent common, indoor, residential home settings, while the bakery scene represents more of a typical commercial business.
We use the 21 objects from YCB~\cite{calli2015ycb} that are included in the {\sc Falling Things} dataset. 

For each stereo frame captured, either a single object was selected at random, or 4 to 10 objects were drawn. Parameters we randomized include object pose (random rotation and placement within a cylinder of 60 radius and 5 height units), 
camera pose, scene lighting (hue, intensity, additional point sources), quantity of objects for mixed captures, and quantity of flying distractors (where present).
Similar to {\sc Falling Things}, the objects of interest follow physics simulation with gravity enabled.
The lighting of the scene comes from a point light that is placed 50cm above each scene point with brightness varying from 5,000--20,000 units and a randomized color hue.
When \emph{flying distractors} are enabled, we select 50--100 from 10 possible meshes (e.g., cubes, cylinders, cones) which appear at random locations with random levels of visibility. Each flying distractor uses one of 7 available material textures at random.

We provide labels as JSON files (two per stereoscopic view pair) including information such as camera location/pose and object location, pose, visibility, and bounding boxes (world space/screen space). Labels are affiliated to their matching images using their filename prefix. 
The dataset is broken down into captures containing single objects and multiple (mixed) objects, as well as with and without distractors. For each of these 4 variations, we capture 2,000 stereo views from each of 18 different viewpoints across 3 scenes (possibly with different camera orientation) for a total of 144k stereo captures, or 288k individual views if stereo pairs are considered separately.
Unlike the {\sc Falling Things} dataset, we did not generate multiple frames during a falling animation, and instead provide individual frames with the randomized parameters and objects.

\begin{figure}
    \centering
    \includegraphics[width=\columnwidth]{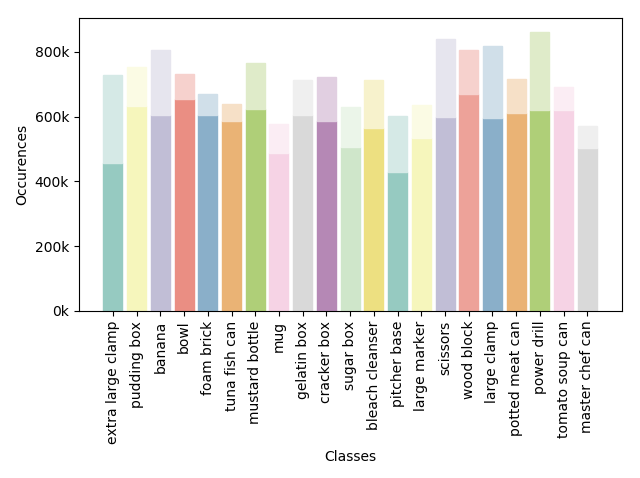}
    \caption{Number of occurrences of each of the 21 YCB objects in our dataset. Light colored bars indicate visibility above 25\% while dark bars indicate greater than 75\% visibility.}
    \label{fig:objdistribution}
\end{figure}

\begin{figure}
    \centering
    \includegraphics[width=\columnwidth]{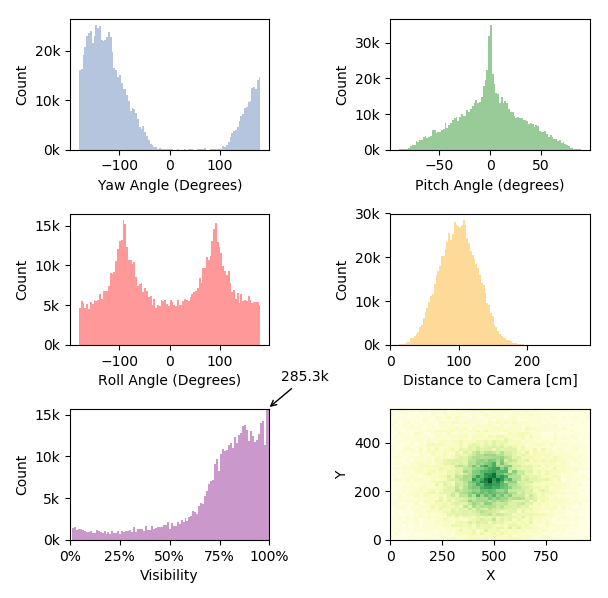}
    \caption{Statistics for the mustard bottle in our dataset. Additional object statistics are available with the dataset.}
    \label{fig:mustardstats}
\end{figure}

\textbf{Statistical analysis:} Figure~\ref{fig:objdistribution} displays the total number of occurrences of each object across our entire dataset. Note that the results provide a variety of high-visibility and low-visibility cases where object occlusion is present. We hope these occlusions can provide meaningful data for training systems to deal with user hand occlusions present during interaction. The amount of occlusion varies based on the relative size of the objects and the other objects and distractors that may also be present.
Figure~\ref{fig:mustardstats} shows the pitch/yaw/roll angles, distance to/from the camera, and screen location heatmap for one object in the 21 chosen from the YCB dataset (mustard bottle).

One additional consideration is that the flying distractors in particular were not carefully controlled to avoid clipping through other objects in the scene. These objects are not intended to directly represent objects that might exist in the real world, and instead provide noise and variation similar to unexpected objects, therefore this clipping should not significantly affect their usefulness beyond the already non-realistic inclusion of them in the first place. It may therefore make sense to use different learning rates when training a network on the part of the dataset with flying distractors than should be used with the rest of the dataset.

\section{Conclusion}
We present a new dataset with distractors created using the NVIDIA NDDS toolkit~\cite{to2018ndds}. This dataset includes 18 view locations from 3 scenes and includes surface normals, flying distractors, and extreme lighting variations. {\sc SIDOD} is more than twice as large as the {\sc Falling Things} dataset and provides significantly more parameter variations.    
We hope that this dataset will enable researchers in the areas of computer vision and robotics to further their goals in training networks more robust to occlusion and unexpected environmental factors.


\section{Acknowledgements}
We would like to acknowledge the many discussions we had during the development of this dataset with our collaborators at NVIDIA including Niveditha Kalavakonda, Turner Whitted, David Luebke, Thang To, Jonathan Tremblay, Omer Shapira, and others. M.B. acknowledges support by DARPA SBIR/STTR.

{\small
\bibliographystyle{ieee_fullname}
\bibliography{egbib}
}

\end{document}